\documentclass[preprint,review,12pt]{elsarticle}

\usepackage{amssymb}

\usepackage[]{hyperref}

\usepackage{colortbl}
\usepackage{color}
\usepackage{tabularray}
\usepackage{multirow}
\usepackage{vcell}
\usepackage{colortbl}

\journal{Pattern Recognition}

\begin{document}

\begin{frontmatter}



\title{Dual-Teacher Ensemble Models with Double-Copy-Paste for 3D Semi-Supervised Medical Image Segmentation}

\author[label1,label2]{Zhan Fa}
\ead{fz1372306571@gmail.com}
\author[label1,label2]{Shumeng Li}
\ead{lism@smail.nju.edu.cn}
\author[label1,label2]{Jian Zhang}
\ead{zhangjian7369@smail.nju.edu.cn}

\author[label3]{Lei Qi}
\ead{qilei@seu.edu.cn}
\author[label4]{Qian Yu}
\ead{yuqian@sdwu.edu.cn}
\author[label1,label2]{Yinghuan Shi\corref{cor1}}
\cortext[cor1]{Corresponding author.}
\ead{syh@nju.edu.cn}

\affiliation[label1]{organization={State Key Laboratory for Novel Software Technology},
             addressline={Nanjing University},
             city={Nanjing},
             country={China}}
             
\affiliation[label2]{organization={National Institute of Healthcare Data Science},
             addressline={Nanjing University},
             city={Nanjing},
             country={China}}
             
\affiliation[label3]{organization={Key Laboratory of New Generation Artificial Intelligence Technology and Its
Interdisciplinary Applications(Ministry of
Education)},
             addressline={Southeast University},
             city={Nanjing},
             country={China}}
    
\affiliation[label4]{organization={School of Data and Computer Science},
             addressline={Shandong Women’s University},
             city={Jinan},
             country={China}}

\begin{abstract}


Semi-supervised learning (SSL) techniques address the high labeling costs in 3D medical image segmentation, with the teacher-student model being a common approach. However, using an exponential moving average (EMA) in single-teacher models may cause coupling issue, where the weights of the student and teacher models become similar, limiting the teacher's ability to provide additional knowledge for the student. Dual-teacher models were introduced to address this problem but often neglected the importance of  maintaining teacher model diversity, leading to coupling issue among teachers. To address the coupling issue, we incorporate a double-copy-paste (DCP) technique to enhance the diversity among the teachers. Additionally, we introduce the Staged Selective Ensemble (SSE) module, which selects different ensemble methods based on the characteristics of the samples and enables more accurate segmentation of label boundaries, thereby improving the quality of pseudo-labels. Experimental results demonstrate the effectiveness of our proposed method in 3D medical image segmentation tasks. Here is the code link: \textcolor{red}{\href{https://github.com/Fazhan-cs/DCP}{https://github.com/Fazhan-cs/DCP}}
\end{abstract}




\end{frontmatter}


\section{Introduction}
Medical image segmentation is a crucial component of computer-aided diagnosis. In recent years, deep learning methods have emerged as the mainstream approach in the field, leading to significant advancements in medical image segmentation \cite{ronneberger2015u}\cite{zhou2023xnet}\cite{Isensee_Jaeger_Kohl_Petersen_Maier-Hein_2021}. However, traditional supervised deep learning methods require a substantial amount of labeled data for training, and pixel-level annotation of medical images is often a time-consuming and challenging task \cite{you2023bootstrapping}\cite{Yan_Liu_Xu_Dong_Li_Shi_Zhang_Dai_2023}. To overcome this problem, semi-supervised medical image segmentation techniques have emerged.

Semi-supervised medical image segmentation leverages a limited amount of labeled data and a large amount of unlabeled data for model training. Recent studies have mainly focused on consistency regularization methods \cite{yu2019uncertainty}\cite{li2020shape}, which encourage models to produce consistent predictions on the same unlabeled inputs, and pseudo-labeling \cite{sohn2020fixmatch}, which involves using reliable pseudo-labels for supervised learning. However, many methods \cite{luo2021semi}\cite{li2020shape} sare based mainly on a single model, which can introduce noise in the training process, leading to low-quality pseudo label, while teacher-student based method may reach a performance bottleneck due to deep coupling  \cite{na2024switching}, which refers to a situation where a single teacher is prone to accumulate errors and the weights of the student and teacher models become similar \cite{arazo2020pseudo}\cite{ke2019dual}, limiting the ability of the teacher model to provide knowledge to the student. Dual-teacher models utilize an ensemble training strategy \cite{liu2022perturbed}\cite{zhao2023alternate}, enhancing model performance by encouraging the dual-teacher to provide additional guidence for the student model. Previous dual-teacher based methods \cite{na2024switching} primarily focus on enhancing the differences between the dual teachers and the student model, this helps prevent the student model from being biased towards a specific direction. However, they overlook the potential guidance that can be provided to the student model by considering the diversity among the dual teachers. To address this issue, we propose a double-copy-paste strategy specifically designed for the dual-teacher models. By incorporating different copy-paste methods \cite{yun2019cutmix}\cite{bai2023bidirectional}, we create diversity in the input data during the training process of the dual teachers, which in turn increase the diversity of the dual teachers themselves. This approach enables the dual-teacher model to benefit not only from ensemble training but also from the different perspectives of the dual teachers, promoting a more robust and effective learning process.

Moreover, previous dual-teacher based ensemble methods \cite{na2024switching}\cite{zhao2023alternate} tend to be fixed during the pseudo-label generation process, and do not take into account the characteristics of the input samples, resulting in poor quality of the generated pseudo-labels. In our work, we propose a novel Staged Selective Ensemble(SSE) module, our proposed method introduces prediction similarity thresholds and allows for flexible selection of ensemble strategies based on the characteristics of the samples. For samples where both dual teachers exhibit high confidence and their predictions are largely consistent, we employ a conservative ensemble strategy to ensure the accuracy and reliability of generating pseudo-labels. In contrast, for samples where there is significant discrepancy in the predictions of the dual teachers, we adopt ensemble strategy with loose limitations to enhance the robustness and adaptability of the pseudo-label generation process. This approach can fully leverage the knowledge of the dual teachers, resulting in the generation of higher-quality pseudo-labels and ultimately improving overall segmentation performance.

Our study aims to explore a semi-supervised medical image segmentation method based on the dual-teacher model. We introduce a novel double-copy-paste technique to enhance input diversity between the dual-teacher. On one hand, this helps capture a more comprehensive range of semantic knowledge. On the other hand, the differences in the training paths taken by the teachers encourage them to learn unique feature representations, which enhances the generalization of the student model. Additionally, we have introduced a new staged selective ensemble module for label generation and employed temporal updates for the the teacher models. Through these approaches, we can effectively utilize limited labeled data and a large amount of unlabeled data while addressing model coupling issues, and generate high-quality pseudo labels, thus enhancing the performance of the model.

Our contributions can be summarized as follows:
\begin{itemize}
    \item We propose a dual-teacher ensemble model with a novel staged ensemble labeling method for medical image segmentation. This approach reduces the coupling issues between the teachers, avoiding the performance bottleneck, and generating high-quality pseudo labels.
    \item We delve into and extend the copy-paste input augmentation method, and propose a novel double-copy-paste technique to enhance the diversity of the dual-teacher model, effectively exploring the potential in leveraging unlabeled data.
    \item Our proposed method is concise and applicable to various datasets and settings. We achieve outstanding results in semi-supervised segmentation tasks on three classic datasets, surpassing the performance of fully supervised methods in certain metrics.

\end{itemize}

The rest of this paper is organized as follows: Section 2 introduces the related work in this paper. Section 3 presents the detailed modules of our proposed method. In Section 4, we demonstrate the experimental details and results of our method on multiple dataset settings, comparing them with state-of-the-art methods. Section 5 provides a relevant analysis of our results, and finally, we conclude this work.

\section{Related Work}
\subsection{Semi-Supervised Medical Image Segmentation}
Semi-supervised learning (SSL) has been widely applied in various computer vision tasks by leveraging both labeled and unlabeled images for training. Consistency learning has been widely used in recent Semi-Supervised Medical Image Segmentation methods \cite{yu2019uncertainty}\cite{li2020shape}\cite{wu2022mutual}\cite{wu2022exploring}. It makes the decision boundary of the learned model located within the low-density region \cite{jiao2023learning}, and aims to enhance the performance of models by encouraging the consistent prediction on unlabeled data under different perturbations. According to the difference of perturbations, consistency learning is divided into three types: input perturbation, feature perturbation, and network perturbation.

SSASNet \cite{li2020shape} utilizes unlabeled data to constrain the geometric shape of segmentation outputs. SS-Net \cite{wu2022exploring} employs a prototype-based strategy to separate features of different classes, promoting category-level separation, and utilizes cross-entropy regularization for model training. DTC \cite{luo2021semi} proposes a dual-task consistency framework by explicitly constructing task-level regularization. It simultaneously performs two tasks during training and enhances consistency between the two tasks through task-level regularization. UAMT \cite{yu2019uncertainty} introduces an uncertainty-aware training scheme to alleviate bias, which helps improve the model's performance on real-world data. MC-Net+ \cite{wu2022mutual} leverages uncertain predictions and applies a mutual consistency constraint to improve performance. CauSSL \cite{miao2023caussl} proposes a novel causal graph that provides a theoretical foundation for mainstream semi-supervised segmentation methods.

Most of these methods employ consistency regularization techniques and introduce additional tasks or networks to provide constraints or perturbations for performance improvement and model robustness. However, these methods primarily rely on a single model or a teacher-student model. On one hand, the training process can be influenced by noise, leading to biases in segmentation tasks. On the other hand, the coupling of weights in the teacher-student model can create performance bottlenecks.

\subsection{Copy-Paste Strategies}
The copy-paste technique involves copying and pasting cropped regions from one image to another. 
Classic works in this field include Mixup \cite{zhang2017mixup}, which employs linear interpolation to blend entire images, and CutMix \cite{yun2019cutmix}, involving the cutting of a portion of one image and pasting it onto another. These techniques have been popular in various domains in recent years. CP2 \cite{wang2022cp} introduces Copy-Paste Contrastive Pretraining, a pixel-wise contrastive learning method that improves transfer performance in dense prediction tasks by incorporating both image- and pixel-level representation learning. UCC  \cite{fan2022ucc}is a novel framework for semi-supervised semantic segmentation that incorporates uncertainty modeling, it selectively copies pixels from classes with low confidence scores and treats them as foreground, enhancing the pseudo labels. BCP \cite{bai2023bidirectional} utilizes a Bidirectional Copy-Paste strategy to jointly process labeled and unlabeled data, encouraging prediction consistency between the teacher and student models to reduce empirical distribution mismatch. 

Many of these studies primarily employ the copy-paste strategy as a form of data augmentation, operating only on a small portion of the image while disregarding the background that has not been copied and pasted. However, by combining different copy-paste methods, we can not only make comprehensive use of the image semantic information, but also create diversity in input data to introduce perturbation at  data level, enabling the model to acquire a broader range of knowledge.

\subsection{Multi-Model Segmentation Strategies}
Multi-Model Segmentation Strategies mainly focuses on improvements to the teacher-student model and includes co-training \cite{chen2021semi}\cite{luo2022semi}\cite{zhao2023rethinking}\cite{huang2023complementary} and ensemble methods  \cite{na2024switching}\cite{huo2021atso}\cite{liu2022perturbed}\cite{zhao2023alternate} : The co-training framework aims to introduce another student model and encourages mutual supervision between the two models. Ensemble method requires the joint utilization of predictions from both teacher models to generate high-quality pseudo-labels.

The MC-net+ \cite{wu2022mutual} model consists of a shared encoder and multiple slightly different decoders (employing different upsampling strategies) that leverage unlabeled data and address the ambiguous segmentation. CC-Net \cite{huang2023complementary} incorporates a complementary symmetric structure with a main model and two auxiliary models, where the complementary models perturb each other, promoting complementary consistency. The complementary information obtained from the two auxiliary models helps the main model effectively focus on the ambiguous regions, and the enforced consistency between the models facilitates obtaining low uncertainty decision boundaries. DPMS \cite{zhao2023rethinking} enhances segmentation performance through data perturbation and stabilizing the dual-student models. AD-MT \cite{zhao2023alternate} comprises a single student model and two non-trainable teacher models, which are alternately and randomly momentum-updated using an entropy-based ensemble strategy, it encourages the model to learn from consistent and conflicting predictions between the teachers.

However, this method suffers from performance degradation due to model's lack of diversity. They prioritize the differences between the student and teacher models while overlooking the diversity among the dual-teacher models themselves, and the ensemble method they employed are relatively fixed and do not take into account the characteristics of the input samples.

\section{Method}
In this section, we will provide the overview of our overall approach and the detail of specific modules, including the setup of the Dual-Teacher model, the Double-Copy-Paste technique, the Staged Selective Ensemble module, mask control, and other relevant configurations.

\subsection{Overview}
We define 3D medical images as $X\in \mathbb{R}^{W\times H\times L}$($W$: width, $H$: height, $L$: length), where the task of semi-supervised medical image segmentation is to predict a label from the range of 0 to $K$ for each pixel in the 3D image. Here, $K$ represents the total number of classes. Our training dataset $D$ consists of $N$ labeled data and $M$ unlabeled data, with $N < M$, denoted as $D^l$ and $D^u$, respectively.

\begin{figure}[htbp]
  \centering
  \includegraphics[width=0.8\textwidth]{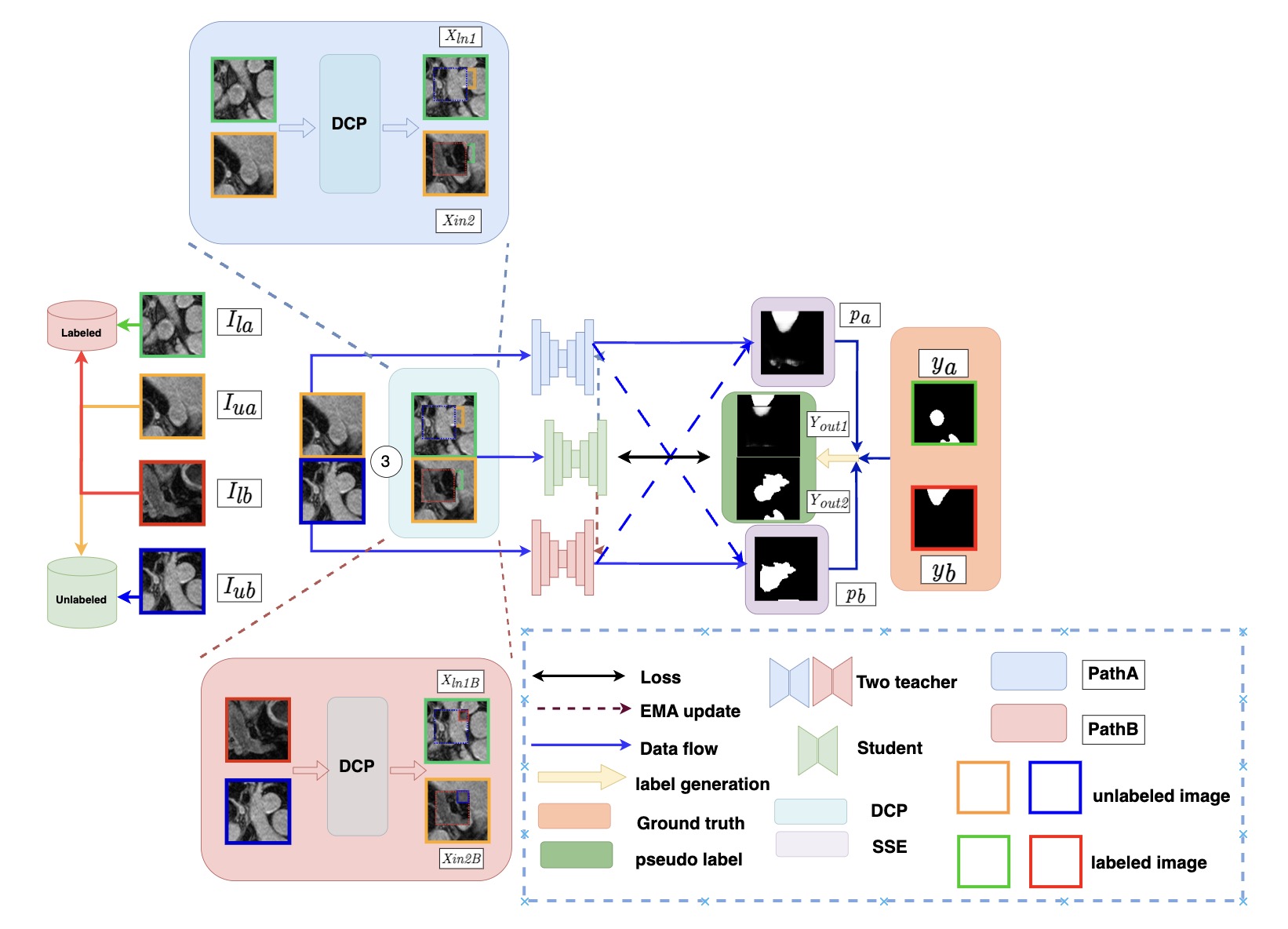}
  \caption{  Illustration of the overall workflow of our method. The images processed through the DCP operation are fed into the student model. Unlabeled images are fed into the teacher models to generate pseudo-labels through the SSE module, and then they are copy-pasted with the ground truth to guide the student model. The weight parameters are updated accordingly, and the student model uses Exponential Moving Average (EMA) to update teacher model based on the current path(A or B).}
  \label{fig:overview}
\end{figure}

In our main dual-teacher training, the overall training process is illustrated in Fig.\ref{fig:overview}. In each iteration, two image groups, $I_{la}$, $I_{ua}$, $I_{lb}$, and $I_{ub}$, are randomly selected from $D^l$ and $D^u$, respectively, with the labeled dataset containing the corresponding ground truth labels. Based on the random selection, the next step is to determine which copy-paste path to choose for the current step, illustrated here using path A (blue background) as an example.
\par
As shown in Fig.\ref{fig:DCP} \textcircled{\small{1}}. Firstly, one image group in Path A, which contains two images $I_{la}$, $I_{ua}$, is subjected to the classic cutmix operation for the first step of copy-paste.
The results are then subjected to a second round of copy-paste with another image group $I_{lb}$, $I_{ub}$ (Fig.\ref{fig:DCP} \textcircled{\small{2}}).
The two processed images $X_{in1}$ $X_{in2}$ are then fed into the student model(Fig.\ref{fig:overview} \textcircled{\small{3}}). The segmentation predictions are used to calculate the semi-supervised loss with generated pseudo labels (green background), which integrate the predictions of the dual-teacher network and the ground truth labels. The student model parameters are updated accordingly.

\subsection{Module Detail}
\subsubsection{Dual-Teacher and Training Strategy}
In our dual-teacher framework, we employ two teacher models, $F_{t1}(\theta_{t1}) $ and $F_{t2}(\theta_{t2}) $, along with a student model, $F_{s}(\theta_{s}) $, where $\theta_t$ and $\theta_s$ represent the model parameters. Our training strategy consists of several parts. Initially, we pretrain the student model using only labeled data . Subsequently, we initialize the dual-teacher models with the pretrained model parameters. Following that, we generate pseudo labels for the unlabeled images and use the integrated pseudo labels to calculate the loss by comparing them to the outputs of the student model. Within each iteration, we optimize the student model parameters, $\theta_s$. Finally, we update the teacher model parameters  $\theta_t$ , by employing the Exponential Moving Average (EMA) of the student model parameters.

\subsubsection{Copy-Paste Augmented Pretraining}
Inspired by \cite{ghiasi2021simple}\cite{bai2023bidirectional}, We initially applied a basic Copy-Paste augmentation technique to the labeled data for pretraining the supervised model. This process enables us to effectively leverage the labeled data, thereby equipping the student and teacher models with initial weights and a certain level of performance for subsequent training.

\begin{figure}[htbp]
  \centering
  \includegraphics[width=0.9\textwidth]{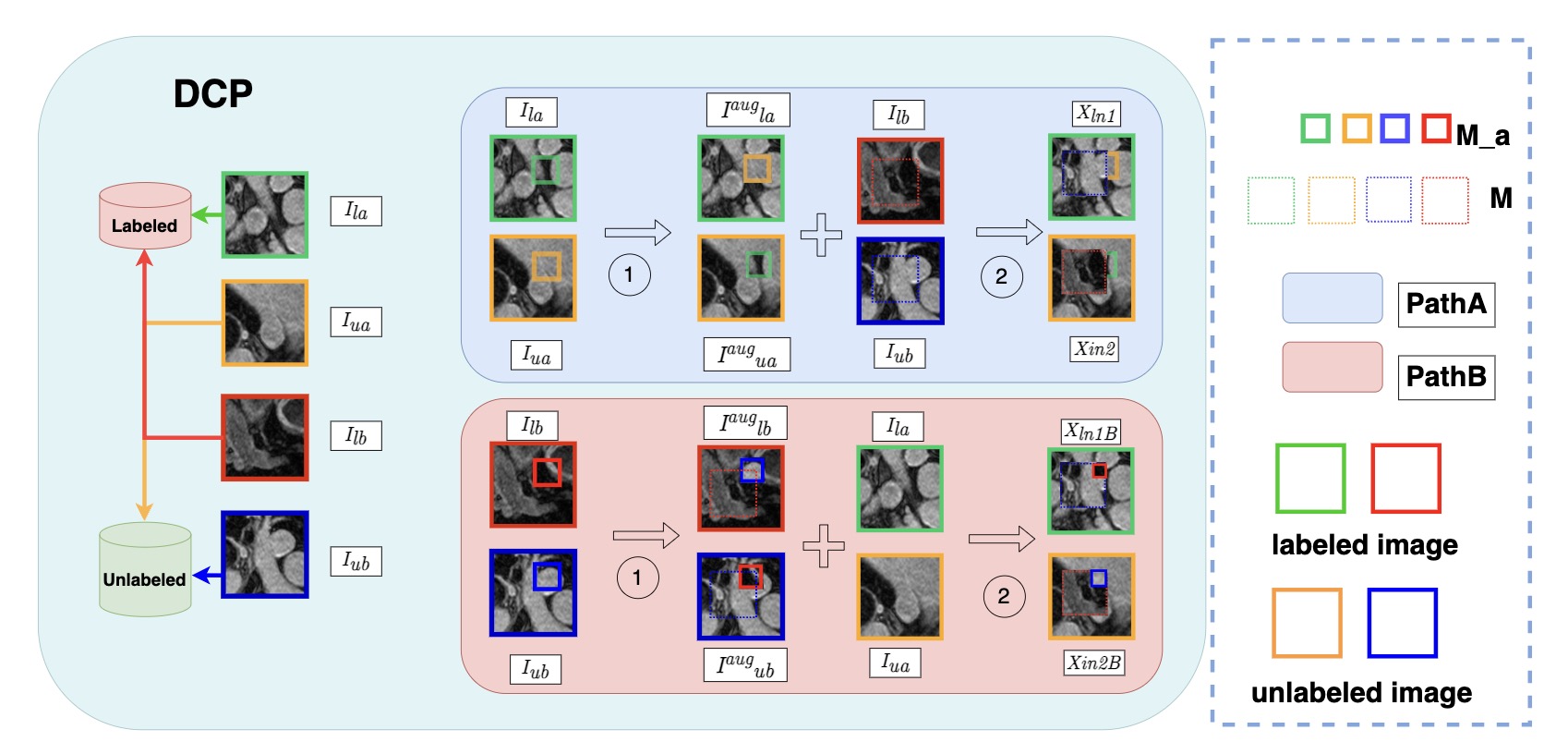}
  \caption{  Illustration of the DCP module. Different copy-paste strategies are devised for different path (A or B), comprising Step 1  \textcircled{\small{1}}and Step  2 \textcircled{\small{2}}. Step 1 is applied within the same pair of images, while Step 2 is applied between the two pairs of images. Blocks of the same color indicate same images.}
  \label{fig:DCP}
\end{figure}

\subsubsection{Double-Copy-Paste Strategy}
To generate augmented input images specifically designed for the dual-teacher models, reduce their coupling with the student model and increase the diversity between the dual teachers, we divide the augmentation process into two steps and introduce the Double-Copy-Paste strategy, abbreviated as DCP. Based on random selection, each path is chosen with a probability of 0.5, denoted as Path A and Path B. For step 1 in Path A, as shown in Fig.\ref{fig:DCP} \textcircled{\small{1}} (blue background), we need to generate a zero-centered mask, $M_a\in \{0, 1\}^{W\times H\times L}$, to perform basic cutmix operation on $I_{la}$ and $I_{ua}$, resulting in $I^{aug}_{la}$ and $I^{aug}_{ua}$:
\begin{equation} I^{aug}_{la} = I_{la}  \odot M_a + I_{ua}  \odot (\mathbf{1} - M_a),  \label{eq:1}\end{equation}
\begin{equation} I^{aug}_{ua} = I_{ua}  \odot M_a + I_{la}  \odot (\mathbf{1} - M_a).  \label{eq:2}\end{equation}

Subsequently, for step 2 of Path A, we generate another zero-centered mask, $M$, as depicted in Fig.\ref{fig:overview} \textcircled{\small{2}}  (blue background). We then perform bidirectional Copy-Paste \cite{bai2023bidirectional} on $I^{aug}_{la}$ and $I^{aug}_{ua}$, $I_{lb}$ and $I_{ub}$ :
\begin{equation} X_{in1} = I^{aug}_{la} \odot M + I_{ub} \odot (\mathbf{1} - M),  \label{eq:3}\end{equation}
\begin{equation} X_{in2} = I^{aug}_{ua} \odot M + I_{lb} \odot (\mathbf{1} - M).  \label{eq:4}\end{equation}

Here, $M$ and $M_a$ represent cropping a cube within the original input. $\odot$ denotes element-wise multiplication, $\mathbf{1}$ represents $\{1\}^{WxHxL}$.

For the other Path B (red background), in step \textcircled{\small{1}}, we generate $M_b$ to obtain $I^{aug}_{lb}$ and $I^{aug}_{ub}$. The only difference is that in step \textcircled{\small{2}}, we still crop the image group $a$ as the masked part for Path B, i.e., $ X_{in1_B}$ and $X_{in2_B}$ for Path B:

\begin{equation} X_{in1_B} = I_{la} \odot M + I^{aug}_{ub} \odot (\mathbf{1} - M),  \label{eq:5}\end{equation}
\begin{equation} X_{in2_B} = I_{ua} \odot M + I^{aug}_{lb} \odot (\mathbf{1} - M).  \label{eq:6}\end{equation}

These serve as the data augmentation preprocessing steps for subsequent training.

\begin{figure}[htbp]
  \centering
  \includegraphics[width=0.5\textwidth]{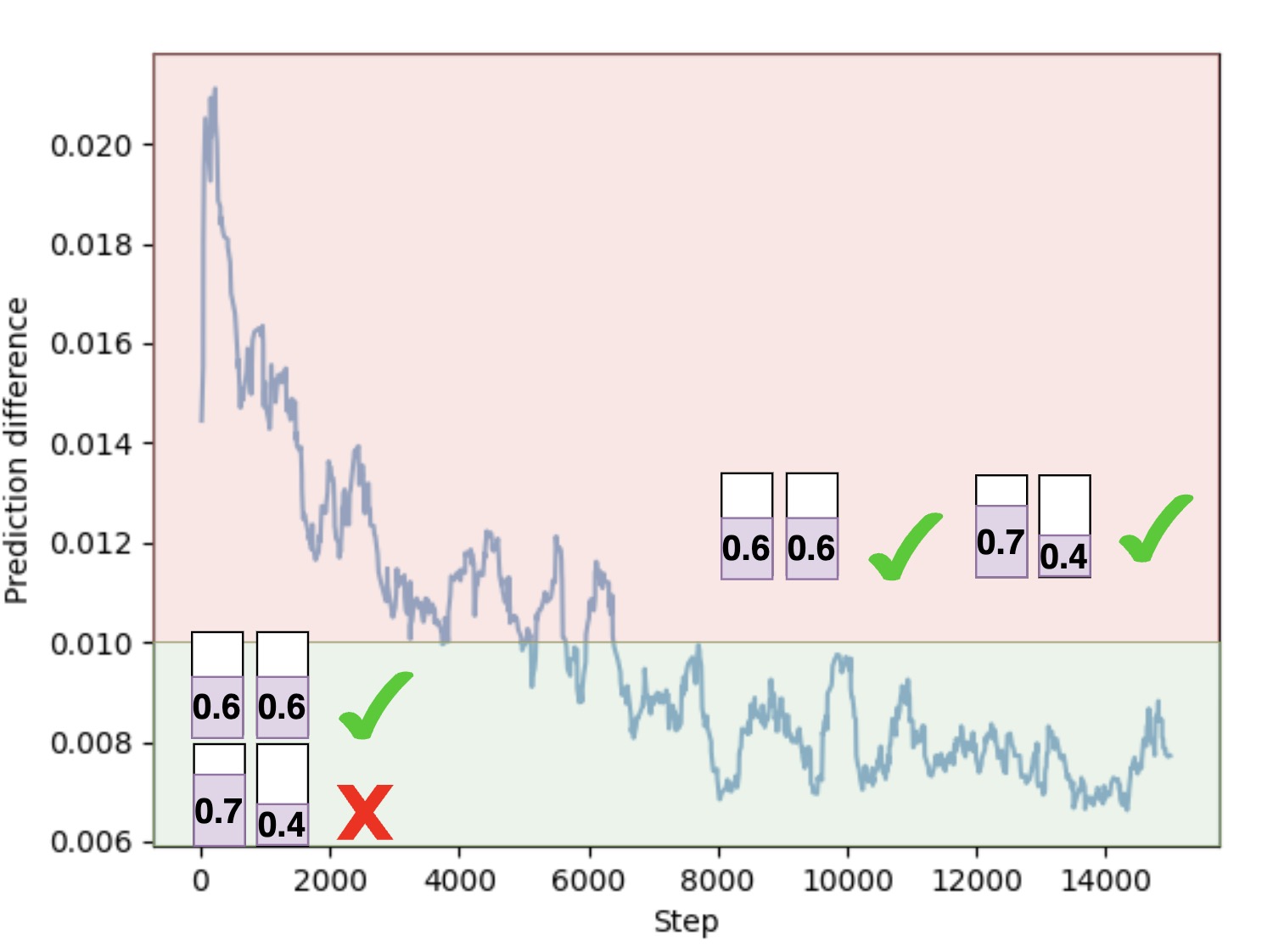}
  \caption{  Illustration of the SSE module. A prediction similarity threshold is set, where samples' prediction with a similarity above the threshold are considered difficult samples, a ensemble method with loose constraints is employed for these samples. Conversely, samples with a similarity below the threshold are considered easy samples, and a strict foreground selection strategy is applied to them.}
  \label{fig:ensemble}
\end{figure}

\subsubsection{Staged Selective Ensemble}

To fully leverage the advantages of the dual-teacher models in our method, we have innovatively designed an ensemble method specifically tailored for the dual-teacher model. We name this method Staged Selective Ensemble (SSE) module. Through experimental observations, we found that the similarity between predictions from the teacher models, calculated using the L1 distance, gradually decreases as the training progresses. This indicates that the teachers gradually reach a consensus in their predictions.

In our approach, as is shown in Fig.\ref{fig:ensemble}(The similarity curve was smoothly processed by EMA), we introduce a threshold to determine the level of dissimilarity between the teachers' predictions. When the dissimilarity between the dual-teacher predictions exceeds the threshold, it indicates that there is still disagreement in the segmentation predictions of the dual teachers. We consider such images as challenging samples and adopt ensemble strategy with loose constraints to select more pixels as foreground for guidance. Specifically, we consider the portions where the sum of the dual-teacher prediction probabilities is greater than 1 as foreground. When the dissimilarity between the dual-teacher predictions is below the threshold, it suggests that the dual teachers have reached a consensus on the sample to some extend. We consider such samples as easy samples and use an ensemble method with stricter constraints to determine more precise label boundaries. In this case, both of the dual-teacher prediction probabilities need to be greater than 0.5 to be considered as foreground.

By employing this ensemble method, we take into consideration the characteristics of different samples and generate precise pseudo-labels. We will present further ablation experiments in Section 5 to demonstrate the effectiveness of our approach.

\subsection{Supervisory Signals and Loss Function}
To train the student network, unlabeled images, $I_{ua}$ and $I_{ub}$, are fed into the teacher networks, and both teachers output predictions for both images. By applying the Staged Selective Ensemble method, we obtain the pseudo labels for the dual-teacher models:\quad $PL_a = \phi (R^{T1}_a,R^{T2}_a)$\quad, 
$PL_b = \phi (R^{T1}_b,R^{T2}_b)$. Here, $\phi$ represents our SSE method, $R^{T1}_a$ and $R^{T1}_b$ represent the predictions of Teacher 1 for the unlabeled images $I_{ua}$ and $I_{ub}$ respectively. By selecting the maximum connected component for each pseudo-label, $PL_a$ and $PL_b$, we obtain the final pseudo-labels, $p_a$ and $p_b$, effectively removing outlier voxels.

Subsequently, similar to the Double-Copy-Paste operations for the input discussed in 3.2.3, we perform similar Copy-Paste operations on the generated label. For the ground truth $y_a$, $y_b$ and generated $p_a$, and $p_b$, the processing of final pseudo labels for Path A is as follows:
\begin{equation} lab^{aug}_{la} = y_a  \odot M_a + p_a  \odot (1 - M_a), \label{eq:9}\end{equation}
\begin{equation} lab^{aug}_{ua} = p_a  \odot M_a + y_a  \odot (1 - M_a), \label{eq:10}\end{equation}
\begin{equation} Y_{out1} = lab^{aug}_{la} \odot M + p_b \odot (1 - M), \label{eq:11}\end{equation}
\begin{equation}Y_{out2} = lab^{aug}_{ua} \odot M + y_b \odot (1 - M). \label{eq:12}\end{equation}

Here, $Y_{out1}$ and $Y_{out2}$ refers to the pseudo-labels corresponding to the $X_{in1}$ and $X_{in2}$ of the student model. Based on intuitive experience, the ground truth of the labeled image is usually more accurate and convincing than the pseudo-label of the unlabeled image. We use $\alpha$ to control the contribution of unlabeled image pixels to the loss function. Unlike previous research, the loss masks generated by different training paths need to be adjusted. Specifically, the loss mask for labeled data, namely $Mask$, is calculated as follows:
\begin{equation}Path A :\qquad Mask = M \wedge M_a , \label{eq:13}\end{equation}
\begin{equation}Path B:\qquad Mask = (1 - M_b) \vee M  . \label{eq:14}\end{equation}

Here, $\wedge$ represents intersection, and $\vee$ represents union.

The loss functions for $X_{in1}$ and $X_{in2}$ are as follows:
\begin{equation}L_{in1} = L_{seg}(Q_{in1}, Y_{out1}) \odot Mask + \alpha \cdot L_{seg}(Q_{in1}, Y_{out1}) \odot (1 - Mask), \label{eq:15}\end{equation}
\begin{equation}L_{in2} = L_{seg}(Q_{in2}, Y_{out2}) \odot (1 - Mask) + \alpha \cdot  L_{seg}(Q_{in2}, Y_{out2}) \odot Mask. \label{eq:16}\end{equation}

Here, $\alpha$ is a weight parameter that controls the loss weight between ground truth labels and pseudo-labels. $L_{seg}$ is a linear combination of the Dice loss and cross-entropy loss. $Q_{in1}$ and $Q_{in2}$ are generated by the student model and calculated as $Q_{in1} = F(X_{in1}; \theta_s)$ and  
$Q_{in2} = F(X_{in2}; \theta_s)$.

In each iteration, we update the parameters $\theta_s$ of the student network using stochastic gradient descent and the loss function:

\begin{equation}L_{all} = L_{in1} + L_{in2}. \label{eq:19}\end{equation}

Next, we update the parameters of the teacher networks, alternating between different teachers based on the chosen path (A or B) during the current training iteration, using exponential moving average:

\begin{equation}Path A: \theta_{t1}(k+1) = \lambda\cdot\theta_{t1}(k) + (1 - \lambda)\cdot\theta_s(k),\label{eq:20}\end{equation}
\begin{equation}Path B: \theta_{t2}(k+1) = \lambda\cdot\theta_{t2}(k) + (1 - \lambda)\cdot\theta_s(k).\label{eq:21}\end{equation}

Here, $\lambda$ is the smoothing coefficient parameter, $k$ is the current iteration number.

\section{Experiment}
\subsection{DataSet}
This experiment mainly utilizes three classic 3D medical image segmentation datasets, covering organ and tumor segmentation.

The \textbf{LA dataset} \cite{xiong2021global} comes from the 2018 Atrial Segmentation Challenge, consisting of 154 3D MRI scans from 60 diagnosed atrial fibrillation patients. Each 3D MRI scan has a resolution of 0.625×0.625×0.625 mm³ with spatial dimensions of 576×576×88 or 640×640×88 pixels. The segmentation labels were manually obtained by three experts and stored in NRRD format. Due to the availability of only 100 annotated images, this dataset follows the setup of previous experiments \cite{wu2022exploring}\cite{bai2023bidirectional}, where the 100 images are divided into 80 for training and 20 for validation. The performance of the proposed method is compared with other methods using the same validation set.

The \textbf{Pancreas-CT dataset} \cite{roth2015deeporgan} consists of 82 3D CT scans from 53 male and 27 female patients. Cases \#25 and \#70 are duplicates of Case \#2 with minor cropping differences and were removed from the dataset. The CT scans have a resolution of 512×512 pixels, and pixel size and slice thickness range from 1.5 to 2.5 millimeters. Following previous studies \cite{shi2021inconsistency}, data augmentation is performed through rotation, scaling, and flipping, and voxel sizes are resampled to a uniform resolution of 1.0×1.0×1.0mm³. Finally, 60 samples are used for training, and the performance is evaluated on the remaining 20 samples.

We also utilize brain tumor segmentation image data from the \textbf{BraTS 2019} \cite{hdtd-5j88-20} challenge. This dataset includes preoperative MRI data (including T1, T1Gd, T2, and T2-FLAIR modalities) from 335 glioma patients. These images are resampled to an isotropic resolution of 1×1×1mm³. Following the same data split and preprocessing procedure as previous work \cite{xu2022all}, 250 samples are used for training, 25 samples for validation, and the remaining 60 samples for testing.
\subsection{Implementation details and evaluation metrics}
Following the previous research \cite{bai2023bidirectional}\cite{luo2021semi}\cite{xu2023ambiguity} on semi-supervised medical image segmentation, for fair comparison, we employed V-Net \cite{milletari2016v} as the backbone for LA and Pancreas-CT dataset experiments, while using 3D-Unet  \cite{cciccek20163d} as the backbone for the Brats 2019 dataset experiment. 

For the \textbf{LA dataset}, data augmentation was performed using rotation and flipping operations, and our model was trained using the SGD optimizer with an initial learning rate of 0.01. The momentum and weight decay were set to 0.9 and 0.0001, respectively. During training, we randomly cropped patches of size 112 × 112 × 80 and set the batch size to 8, including four labeled patches and four unlabeled patches. The pre-training and main training iterations were set to 2k and 15k, respectively.

For the \textbf{Pancreas dataset}, data augmentation was performed using rotation, scaling, and flipping operations, and the model was trained using the SGD optimizer with an initial learning rate of 0.01. The momentum and weight decay were set to 0.9 and 0.0001, respectively. We randomly cropped input patches of size 96 × 96 × 96, and the batch size, pre-training iterations, and main training iterations were set to 8, 100, and 500, respectively.

For the \textbf{Brats 2019 dataset}, we randomly cropped patches of size 96 × 96 × 96 voxels as input and utilized a sliding window strategy with a stride of 64 × 64 × 64 voxels for prediction. The number of pre-training iterations was set to 8000, and the maximum iterations for main training were set to 20,000. The model was initialized using Kaiming initialization.

In contrast to previous studies \cite{bai2023bidirectional} that use different optimizers such as SGD and Adam for different datasets, we used the SGD optimizer for all our experiments and omitted the manual learning rate adjustment step. The parameters of the two teacher models were updated using exponential moving average with a decay rate of 0.99. The threshold for path selection in the dual-teacher framework was set to 0.5. The size ratios for the first and second copy-paste operations were set to 1/3 and 2/3 of the original size, respectively. The similarity thresholds for the dual-teacher were set to 0.01, 0.001, and 0.01 for the LA, Pancreas-CT, and Brats 2019 datasets, respectively.

For fair comparison, We adopt four metrics for  evaluation, including Dice score, Jaccard score, average surface distance (ASD) and 95\% Hausdorff distance (95HD). We implemented and trained our method using PyTorch on an NVIDIA GeForce RTX 2080TI GPU.

\subsection{Comparison with state-of-the-art methods}
We compared our method with various semi-supervised medical image segmentation methods, including UA-MT \cite{yu2019uncertainty}, DTC \cite{luo2021semi}, MC-Net+ \cite{wu2022mutual}, SS-Net \cite{li2020shape}, CC-net \cite{huang2023complementary},  BCP \cite{bai2023bidirectional}, URPC \cite{luo2021efficient}, CauSSL \cite{miao2023caussl}, GenericSSL \cite{wang2024towards}, CAML \cite{gao2023correlation}, AC-MT \cite{xu2023ambiguity}. 

On the LA dataset, We conducted semi-supervised experiments with different labeled data ratios (i.e., 5\%, 10\%, 20\%). As shown in Table.\ref{tab:1}, our method achieves the best performance at 5\%, 10\%, and 20\% labeled images across all four evaluation metrics, outperforming other competitors, particularly surpassing the upper bound of the fully supervised method (Vnet) on certain metrics.

\begin{table}
\centering
\setlength{\tabcolsep}{1mm}
\caption{
Comparison of metrics between the proposed method and other semi-supervised segmentation methods on the \textbf{LA} dataset with varying label ratios.}
\arrayrulecolor{black}
\scalebox{0.6}{
\begin{tabular}{c|c|c c|c c c c}
\noalign{\smallskip}
\hline
\noalign{\smallskip}
\multirow{2}{*}{Method }  & \multirow{2}{*}{Venue }  & \multicolumn{2}{c}{Scans used}                      & \multicolumn{4}{c}{Metrics}                                                                                                                                                                                  \\ 
\cline{3-8}
~  &      & labeled            & unlabeled                     & Dice↑          & Jaccard↑       & 95HD↓                                                                        & ASD↓                                                                                                              \\ 
\noalign{\smallskip}
\hline
\noalign{\smallskip}
Vnet \cite{milletari2016v}      &  -         & \multicolumn{2}{l!{\color{black}\vrule}}{~~~~80(all) ~~~~~~~~~~~~~~~~0}           & 91.47          & 84.36          & 5.48          & 1.51           \\ 
\noalign{\smallskip}
\hline
\noalign{\smallskip}
UA-MT \cite{yu2019uncertainty}  & MICCAI'19                   & \multirow{7}{*}{~~~4(5\%) ~ ~ ~ ~}   & \multirow{7}{*}{~~76(95\%) ~ ~ ~ ~} & 82.26          & 70.98          & 13.71         & 3.82           \\ 

DTC \cite{luo2021semi}      & AAAI'21                 &                                     &                                     & 81.25          & 69.33          & 14.90         & 3.99           \\ 

SS-Net \cite{wu2022exploring}   & MICCAI'22                 &                                     &                                     & 86.33          & 76.15          & 9.97          & 2.31           \\ 

BCP \cite{bai2023bidirectional}   & CVPR'23                    &                                     &                                     & 88.02          & 78.72          & 7.90          & 2.15           \\ 
CAML \cite{gao2023correlation}         & MICCAI'23              &                                     &                                     & 87.34          & 77.65         & 9.76          & 2.49           \\ 
GenericSSL \cite{wang2024towards}       & NeurIPS'23          
&                                     &                                     & 89.93          & 81.85         & \textbf{5.25}          & 1.86           \\ 

Ours          & -            &                                     &                                     & \textbf{90.12} & \textbf{82.04} & 5.55 & \textbf{1.77}  \\ 
\noalign{\smallskip}
\hline
\noalign{\smallskip}
UA-MT \cite{yu2019uncertainty}      & MICCAI'19               & \multirow{7}{*}{~~~8(10\%) ~ ~ ~ ~}  & \multirow{7}{*}{~~72(90\%) ~ ~ ~ ~} & 87.79          & 78.39          & 8.68          & 2.12           \\ 

DTC \cite{luo2021semi}         & AAAI'21              &                                     &                                     & 87.51          & 78.17          & 8.23          & 2.36           \\ 

CC-net \cite{huang2023complementary}    & CIBM'23                &                                     &                                     & 89.42          & 80.95          & 7.37          & 2.17           \\ 

BCP \cite{bai2023bidirectional}         & CVPR'23              &                                     &                                     & 89.62          & 81.31          & 6.81          & 1.76           \\ 
CAML \cite{gao2023correlation}         & MICCAI'23              &                                     &                                     & 89.62          & 81.28          & 8.76          & 2.02           \\ 

GenericSSL \cite{wang2024towards}       & NeurIPS'23          
&                                     &                                     & 90.31          & 82.40         & 5.55          & 1.64         \\ 

Ours        & -              &                                     &                                     & \textbf{91.18} & \textbf{83.84} & \textbf{5.00} & \textbf{1.61}  \\ 
\noalign{\smallskip}
\hline
\noalign{\smallskip}
UA-MT \cite{yu2019uncertainty}    & MICCAI'19                 & \multirow{7}{*}{~~~16(20\%) ~ ~ ~ ~} & \multirow{7}{*}{~~64(80\%) ~ ~ ~ ~} & 88.74          & 79.94          & 8.39          & 2.32           \\ 

DTC \cite{luo2021semi}       & AAAI'21                &                                     &                                     & 89.52          & 81.22          & 7.07          & 1.96           \\ 

CC-net \cite{huang2023complementary}    & CIBM'23                &                                     &                                     & 91.14          & 83.79          & 5.74          & 1.57           \\ 

MC-Net+ \cite{wu2022mutual}   & MICCAI'22                &                                     &                                     & 91.07          & 83.67          & 5.84          & 1.67           \\ 
CAML \cite{gao2023correlation}         & MICCAI'23              &                                     &                                     & 90.78          & 83.19          & 6.11          & 1.68           \\ 
BCP \cite{bai2023bidirectional}  & CVPR'23              &                                     &                                     & 90.81          & 83.24          & 6.12          & 1.63           \\ 

Ours     & -                 &                                     &                                     & \textbf{91.82} & \textbf{84.92} & \textbf{5.11} & \textbf{1.50}  \\
\noalign{\smallskip}
\hline
\noalign{\smallskip}
\end{tabular}
}
\arrayrulecolor{black}
\label{tab:1}
\end{table}

On the Pancreas-CT dataset, the results of the Pancreas dataset are reported in Table.\ref{tab:2}. As observed from the table, our method demonstrates excellent segmentation performance. We achieve significant performance improvements in both the 10\% and 20\% labeled data scenarios. Furthermore, our method surpasses the upper bound of fully supervised methods on certain metrics.

\begin{table}
\centering
\caption{
Comparison of metrics between the proposed method and other semi-supervised segmentation methods on the \textbf{Pancreas} dataset with varying label ratios.}
\arrayrulecolor{black}
\setlength{\tabcolsep}{1mm}
\scalebox{0.6}{
\begin{tabular}{c|c|c c|c c c c}

\noalign{\smallskip}
\hline
\noalign{\smallskip}
\multirow{2}{*}{Method }  & \multirow{2}{*}{Venue }  & \multicolumn{2}{c}{Scans used}                      & \multicolumn{4}{c}{Metrics}                                                                                                                                                                                  \\ 
\cline{3-8}
~  &      & labeled            & unlabeled                     & Dice↑          & Jaccard↑       & 95HD↓                                                                        & ASD↓                                                                                                              \\ 
\noalign{\smallskip}
\hline
\noalign{\smallskip}
Vnet \cite{milletari2016v}  & -  & \multicolumn{2}{l!{\color{black}\vrule}}{~~~~62(all) ~~~~~~~~~~~~~~~~~0}           & 83.48          & 71.99          & 4.44                                                                         & 1.26                                                                                                              \\ 
\noalign{\smallskip}
\hline
\noalign{\smallskip}
UA-MT \cite{yu2019uncertainty} & MICCAI'19  & \multirow{7}{*}{~~~~6(10\%) ~ ~ ~ ~}  & \multirow{7}{*}{~~56(90\%) ~ ~ ~ ~} & 66.44          & 52.02          & 17.04                                                                        & 3.03                                                                                                              \\ 

DTC \cite{luo2021semi}  & AAAI'21   &                                     &                                     & 66.58          & 51.79          & 15.46                                                                        & 4.16                                                                                                              \\ 

URPC \cite{luo2021efficient} & MedIA'22   &                                     &                                     & 73.53          & 59.44          & 22.57                                                                        & 7.85                                                                                                              \\ 

MC-Net+ \cite{wu2022mutual} & MedIA'22 &       &        & 70.00  & 55.66     & 16.03            & 3.87 \\ 

BCP \cite{bai2023bidirectional} & CVPR'23 &       &        & 81.73  & 69.37     & 8.93            & 2.64 \\ 

CauSSL \cite{miao2023caussl} & ICCV'23 &       &        & 72.89  & 58.06     & 14.19            & 4.37 \\ 

Ours  & -  &  &  & \textbf{83.16} & \textbf{71.44} & \textbf{8.04}                                                                & \textbf{2.28}                                                                                                     \\ 
\noalign{\smallskip}
\hline
\noalign{\smallskip}
UA-MT \cite{yu2019uncertainty}  & MICCAI'19 & \multirow{7}{*}{~~~~12(20\%) ~ ~ ~ ~} & \multirow{7}{*}{~~50(80\%) ~ ~ ~ ~} & 76.01          & 62.62          & 10.84                                                                        & 2.43                                                                                                              \\ 

DTC \cite{luo2021semi}   & AAAI'21  &                                     &                                     & 76.27          & 62.82          & 8.70                                                                         & 2.20                                                                                                              \\ 
URPC \cite{luo2021efficient} & MedIA'22   &                   &                    & 80.02          & 67.30          & 8.51     & 1.98                   \\ 
MC-Net+ \cite{wu2022mutual} & MedIA'22 &    &  & 79.37   & 66.83  & 8.52  & 1.72\\ 

BCP \cite{bai2023bidirectional}  & CVPR'23   &                                     &                                     & 82.91          & 70.97          & \textbf{6.43}      
& 2.2                                                                                                               \\ 
CauSSL \cite{miao2023caussl} & ICCV'23 &       &        & 80.92  & 68.26    & 8.11          &  \textbf{1.53}\\

Ours & -   &                                     &                                     & \textbf{83.66} & \textbf{72.18} & 7.39 & 2.57  \\
\noalign{\smallskip}
\hline
\noalign{\smallskip}
\end{tabular}
}
\arrayrulecolor{black}
\label{tab:2}
\end{table}

\begin{figure}[htbp]
  \centering
  \includegraphics[width=0.8\textwidth]{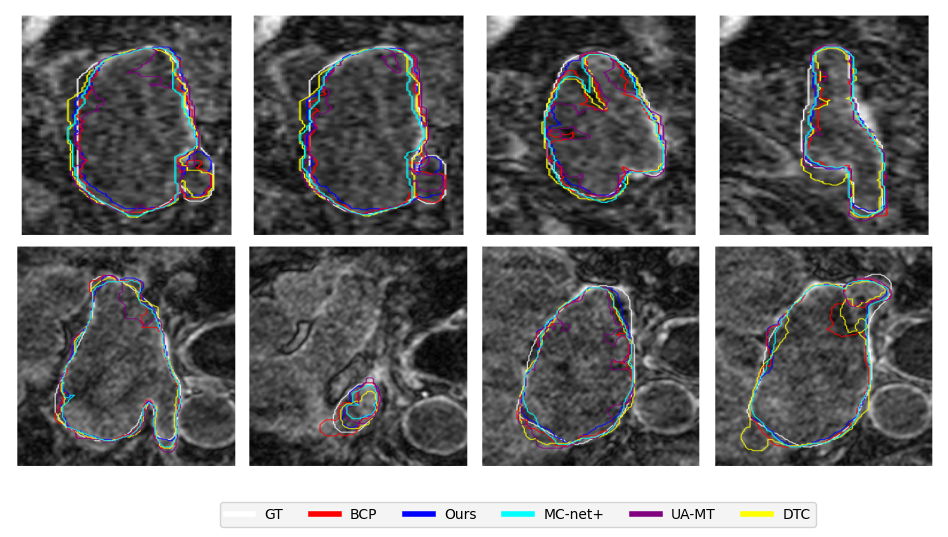}
  \caption{ Visual comparisons of the segmentation results between our method and other methods on multiple medical images on LA 5\%labeled dataset.}
  \label{fig:6}
\end{figure}

On the Brats 2019 dataset, our results, including the metrics for 10\% and 20\% labeled data, are reported in Table.\ref{tab:3}. From the table, it can be observed that our method outperforms all other semi-supervised methods in terms of Dice and Jaccard metrics. Although the other two metrics did not reach the optimal level, they also exhibited excellent performance improvements.
Fig.\ref{fig:6} presents the segmentation results of different semi-supervised methods on the same image, with different colors indicating different segmentation outcomes. From the figure, it can be observed that our method outperforms the previous methods in terms of boundary delineation and edge alignment.

\begin{table}
\centering
\arrayrulecolor{black}
\caption{
Comparison of metrics between the proposed method and other semi-supervised segmentation methods on the \textbf{Brats 2019} dataset with varying label ratios. Underline indicates suboptimal performance.}
\setlength{\tabcolsep}{1mm}
\scalebox{0.6}{
\begin{tabular}{c|c|c c|c c c c}

\noalign{\smallskip}
\hline
\noalign{\smallskip}
\multirow{2}{*}{Method }  & \multirow{2}{*}{Venue }  & \multicolumn{2}{c}{Scans used}                      & \multicolumn{4}{c}{Metrics}                                                                                                                                                                                  \\ 
\cline{3-8}
~  &      & labeled            & unlabeled                     & Dice↑          & Jaccard↑       & 95HD↓                                                                        & ASD↓                                                                                                              \\ 
\noalign{\smallskip}
\hline
\noalign{\smallskip}
3D-Unet \cite{cciccek20163d}     &     -         & \multicolumn{2}{l!{\color{black}\vrule}}{ ~ 250 (all)~ ~ ~ ~ ~ ~ ~ ~ 0}     & 85.93            & 76.81          & 9.85                                                                                                                                                  & 1.93                                                                                                                                                  \\ 
\noalign{\smallskip}
\hline
\noalign{\smallskip}
UA-MT \cite{yu2019uncertainty}     & MICCAI'19                & \multirow{7}{*}{~~25 (10\%) ~ ~ ~} & \multirow{7}{*}{~~225 (90\%) ~ ~ ~} & 80.72~           & 70.30          & 11.76                                                                                                                                     & 2.72  \\ 

DTC \cite{luo2021semi}    & AAAI'21                   &                                      &                                       & 81.75            & 71.63          & 15.73                                                                                                                                                 & 2.56                                                                                                                                         \\ 

URPC \cite{luo2021efficient}    & MedIA'22                  &                                      &                                       & 82.59            & 72.11          & 13.88                                                                                                                                                 & 3.72                                                                                                                                                  \\ 

BCP \cite{bai2023bidirectional}     & CVPR'23                  &                                      &                                       & 84.68            & 74.60          & 17.48                                                                                                                                                 & 4.97                                                                                                                                                  \\ 
CC-Net \cite{huang2023complementary}     & CIBM'23                  &                                      &                                       & 82.74            & 72.82          & 12.29                                                                                                                                                 & 3.02                                                                                                                                              \\ 
AC-MT \cite{xu2023ambiguity}     &  MedIA'23  &                                      &                                       & 83.77            & 73.96          & \textbf{11.37 }                                                                                                                                              & \textbf{1.93}      
\\
Ours     & -                 &                                      &                                       & \textbf{85.15}   & \textbf{75.30} & 12.46\textbf{}                                                                                                                                        & 3.73\textbf{}                                                                                                                                         \\ 
\noalign{\smallskip}
\hline
\noalign{\smallskip}
UA-MT \cite{yu2019uncertainty}      & MICCAI'19               & \multirow{7}{*}{~~50 (20\%) ~ ~ ~} & \multirow{7}{*}{~~200 (80\%) ~ ~ ~} & 83.12            & 73.01          & 9.87                                                                                                                                     & 2.30                                                                                                                                        \\ 

DTC \cite{luo2021semi}       & AAAI'21                &                                      &                                       & 83.43            & 73.56          & 14.77                                                                                                                                                 & 2.34                                                                                                                                                  \\ 

URPC \cite{luo2021efficient}    &MedIA'22                  &                                      &                                       & 82.93            & 72.57          & 15.93                                                                                                                                                 & 4.19                                                                                                                                                  \\ 

BCP \cite{bai2023bidirectional}     &CVPR'23                  &                                      &                                       & 84.94            & 75.15          & 14.65                                                                                                                                                 & 3.75                                                                                                                                                  \\ 
CC-Net \cite{huang2023complementary}     & CIBM'23                  &                                      &                                       & 83.30            & 73.34          & 10.44                                                                                                                                                 & 2.57                                                                                                                                                  \\ 
AC-MT \cite{xu2023ambiguity}     &  MedIA'23  &                                      &                                       & 84.63            & 74.39          & \textbf{9.50}                                                                                                                                                & \textbf{2.11}                     
\\
Ours      &-                &                                      &                                       & \textbf{85.61} & \textbf{76.00} & 10.37\textbf{}                                                                                                                                        & 2.75\textbf{}                                                                                                                                         \\
\noalign{\smallskip}
\hline
\noalign{\smallskip}
\end{tabular}
}
\arrayrulecolor{black}
\label{tab:3}
\end{table}

\begin{table}
\centering

\arrayrulecolor{black}
\caption{Ablation studies on different module on LA 5\% labeled dataset.
}
\scalebox{0.6}{
\begin{tabular}{c|c|c|c|c|c|c|c|c|c} 
\noalign{\smallskip}
\hline
\noalign{\smallskip}
\multicolumn{6}{c|}{Module~employed}                                                                                                                                                         & \multicolumn{4}{c}{4 labels~(5\%labeled)~~~~~~~}                                                                                                                                                                                          \\ 
\noalign{\smallskip}
\hline
\noalign{\smallskip}

\begin{tabular}[c]{@{}l@{}}Single\\teacher\\PathA\end{tabular} & 
\begin{tabular}[c]{@{}l@{}}Single\\teacher\\PathB\end{tabular} & 
\begin{tabular}[c]{@{}l@{}}Concurrent\\teachers'\\update~\end{tabular} & \begin{tabular}[c]{@{}l@{}}Seperate\\teacher\end{tabular} & \begin{tabular}[c]{@{}l@{}}Ensemble\\teacher\end{tabular} & \begin{tabular}[c]{@{}l@{}}Masked\\loss\end{tabular} & Dice↑                                                                         & Jaccard↑ ~                                                                    & 95HD↓         & ASD↓                                                                          \\ 
\noalign{\smallskip}
\hline
\noalign{\smallskip}
~                                          & ~      & ~          & ~                                                         & ~                                                         & ~                                                    & 88.02\textcolor[rgb]{0.051,0.051,0.051}{}                                     & 78.72\textcolor[rgb]{0.051,0.051,0.051}{}                                     & 7.90          & 2.15\textcolor[rgb]{0.051,0.051,0.051}{}                                      \\ 

\checkmark                                                        & ~       & ~    & ~                                                      & ~                                                         & \checkmark                                                     & 88.36 & 79.29 & 8.03          & 2.57 \\ 
~                                                        & \checkmark       & ~    & ~                                                      & ~                                                         & \checkmark                                                     & 89.20 & 80.58 & 5.83          & 2.06  \\ 
~                                                        & ~                                                         & \checkmark                                                         & ~  & ~ &  \checkmark                                                 & 89.62                                                                         & 81.28                                                                         & 6.87          & 1.95                                                                          \\ 
~                                                        & ~                                                         & ~                                                         & \checkmark  & ~ &  \checkmark                                                 & 88.97                                                                         & 80.21                                                                         & 7.23          & 2.14                                                                          \\ 

~                                                        & ~  & ~  & ~                                                        & \checkmark                                                         & ~                                                    & 89.57                                                                         & 81.17                                                                         & 7.23          & 2.12                                                                          \\ 

~                                                & ~     & ~     & ~                                                         & \checkmark                                                         & \checkmark                                                    & \textbf{90.12}                                                                & \textbf{82.04}                                                                & \textbf{5.55} & \textbf{1.77}                                                                 \\
\noalign{\smallskip}
\hline
\noalign{\smallskip}
\end{tabular}
}
\arrayrulecolor{black}
\label{tab:4}
\end{table}

\section{Further analysis}

\subsection{Abition study on different module}
Table.\ref{tab:4} presents the ablation experiments on the relevant modules, the term "Single teacher" refers to the training approach where only a single copy-paste path (A/B) is utilized. "Concurrent teachers' update" indicates the simultaneous update of two teachers within a single iteration. "Separate teacher" indicates pseudo-labels are generated by one teacher independently. "Ensemble teacher" denotes the utilization of a dual-teacher ensemble method. It can be observed that the dual-teacher's guidance significantly outperforms the single-teacher approach. Additionally, the dual-teacher ensemble labeling method outperforms individual guidance. Furthermore, by incorporating different weighted label masks for different paths, our method achieves the optimal performance.

\begin{figure}[htbp]
  \centering
  \includegraphics[width=0.45\textwidth]{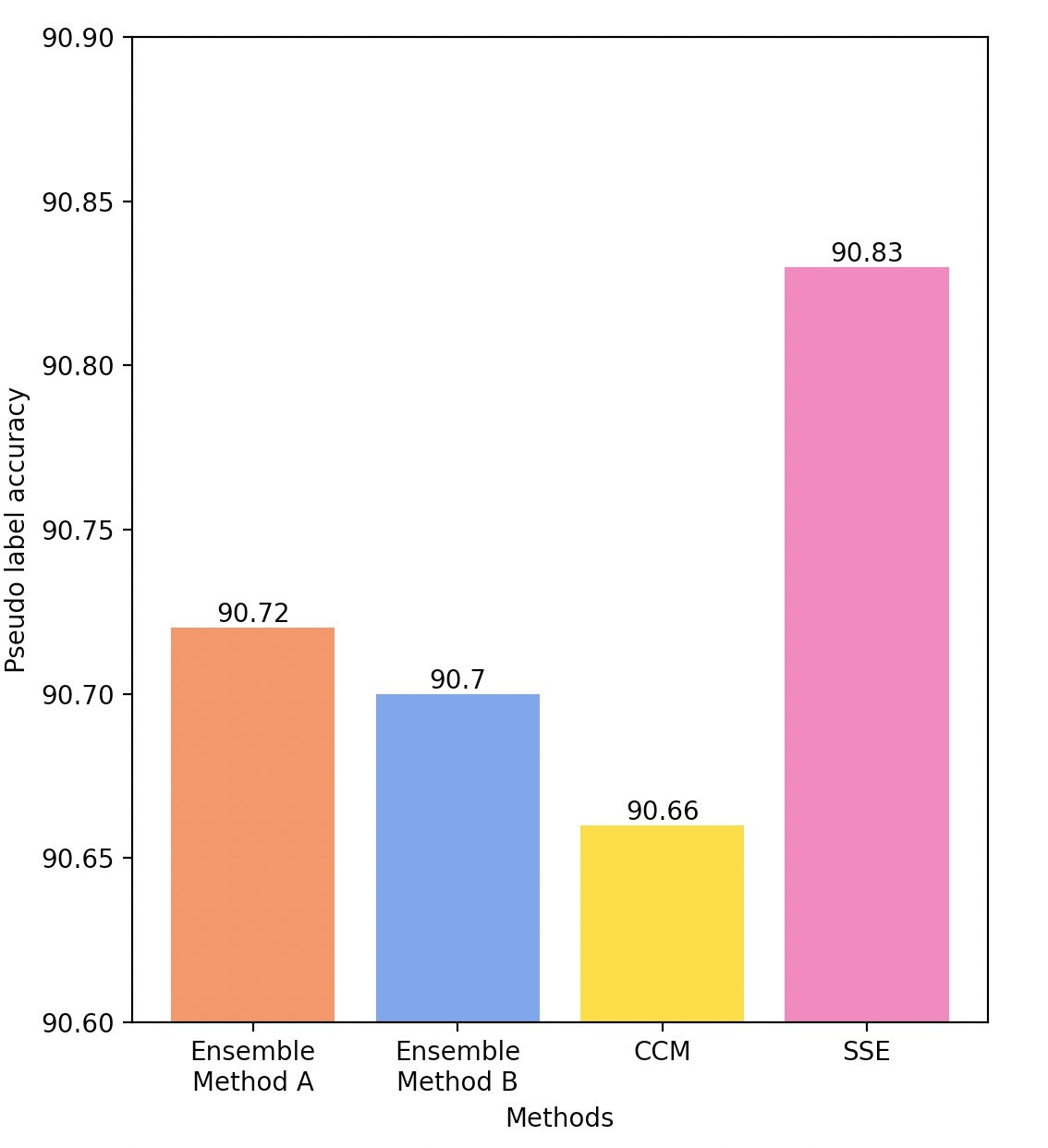}
  \caption{ Comparison of Dice scores for pseudo-label quality obtained using different dual-teacher ensemble methods.}
  \label{fig:plabel}
\end{figure}

\subsection{Abition study on dual-teacher ensemble method}
In order to further investigate the impact of the Staged Selective Ensemble(SSE) module on the model, we designed relevant experiments. From Table.\ref{tab:5}, method A is a classic ensemble method that utilizes the average predictions of the two teachers, it takes the portion where the sum of the probabilities from the dual-teacher predictions is greater than 1 as foreground, while method B represents taking the portion where both of the probabilities from the dual-teacher predictions are greater than 0.5 as foreground. It can be observed that both ensemble methods A and B , which are relatively simple and straightforward, and CCM ensemble method proposed in \cite{zhao2023alternate}, are helpful in improving our model's performance. However, setting a similarity threshold for dual-teacher predictions, thereby providing different levels of strictness for label generation at different stages, leads to our model achieving the best performance. As depicted in Fig.\ref{fig:plabel}, our SSE module generates pseudo-labels that exhibit higher average precision compared to other methods. This provides superior guidance to the model during the training process.

Furthermore, we conducted ablation experiments on three datasets (5\% labeled data) to investigate the setting of the similarity threshold in our SSE module. The results are shown in the Table.\ref{tab:8}, the experimental results show that we can achieve certain performance improvements by setting every threshold introduced in the table. Finally, we selected the hyperparameter recommended in section 3 as the optimal threshold, resulting in the best performance.

\begin{table}
\centering
\arrayrulecolor{black}
\caption{Ablation studies on ensemble methods on LA 5\% labeled dataset.
}
\scalebox{0.8}{
\begin{tabular}{c|c|c|c|c} 
\noalign{\smallskip}
\hline
\noalign{\smallskip}
Settings                               & Dice↑ & Jaccard↑ & 95HD↓ & ASD↓  \\ 
\arrayrulecolor{black}
\noalign{\smallskip}
\hline
\noalign{\smallskip}
Ensemble method A                      & 89.80 & 81.57    & 6.77  & 2.12  \\ 

Ensemble method B                      & 89.72 & 81.48    & 6.47  & 1.83  \\ 
CCM \cite{zhao2023alternate} method                       & 89.85 & 81.63    & 6.00 & 1.86  \\ 

SSE method (this paper) & \textbf{90.12} & \textbf{82.04}    & \textbf{5.55}  & \textbf{1.77}  \\
\noalign{\smallskip}
\hline
\noalign{\smallskip}
\end{tabular}
\arrayrulecolor{black}
\label{tab:5}
}
\end{table}

\begin{table}
\centering
\arrayrulecolor{black}
\caption{Ablation study on similarity threshold on 5\% labeled data
}
\scalebox{0.8}{
\begin{tabular}{c|c|c|c|c|c} 
\noalign{\smallskip}
\hline
\noalign{\smallskip}
\multirow{2}{*}{Dataset}     &  similarity        & \multirow{2}{*}{Dice↑} & \multirow{2}{*}{Jaccard↑} & \multirow{2}{*}{95HD↓} & \multirow{2}{*}{ASD↓}  \\ 
& threshold  & & & &\\
\arrayrulecolor{black}
\noalign{\smallskip}
\hline
\noalign{\smallskip}

\multirow{3}{*}{LA }                  & 0.005  & 89.94 & 81.76    & \textbf{5.27}  & 1.90  \\ 

& 0.01 & \textbf{90.12} & \textbf{82.04}    & 5.55  & \textbf{1.77}  \\
& 0.02 & 89.86 & 81.64   & 5.61  & 1.89  \\

\noalign{\smallskip}
\hline
\noalign{\smallskip}

\multirow{3}{*}{Pancreas }                  &  0.0005 & 82.72 & 70.91    & 9.33  & 2.40  \\ 

& 0.001 & \textbf{83.16} & \textbf{71.44}    & 8.04  & \textbf{2.28}  \\
& 0.002 &82.84 & 70.97    & \textbf{8.03}  & 2.49   \\

\noalign{\smallskip}
\hline
\noalign{\smallskip}

\multirow{3}{*}{Brats }                  & 0.005  & 84.55 & 74.18    & 13.57  &\textbf{ 3.67 } \\ 

& 0.01 &\textbf{85.15}   & \textbf{75.30} & \textbf{12.46}                                                                                                                                        & 3.73                      \\
& 0.02 & 85.01 & 75.15   & 14.67  & 4.17  \\

\noalign{\smallskip}
\hline
\noalign{\smallskip}
\end{tabular}
}
\arrayrulecolor{black}
\label{tab:8}

\end{table}

\begin{figure}[htbp]
  \centering
  \includegraphics[width=1\textwidth]{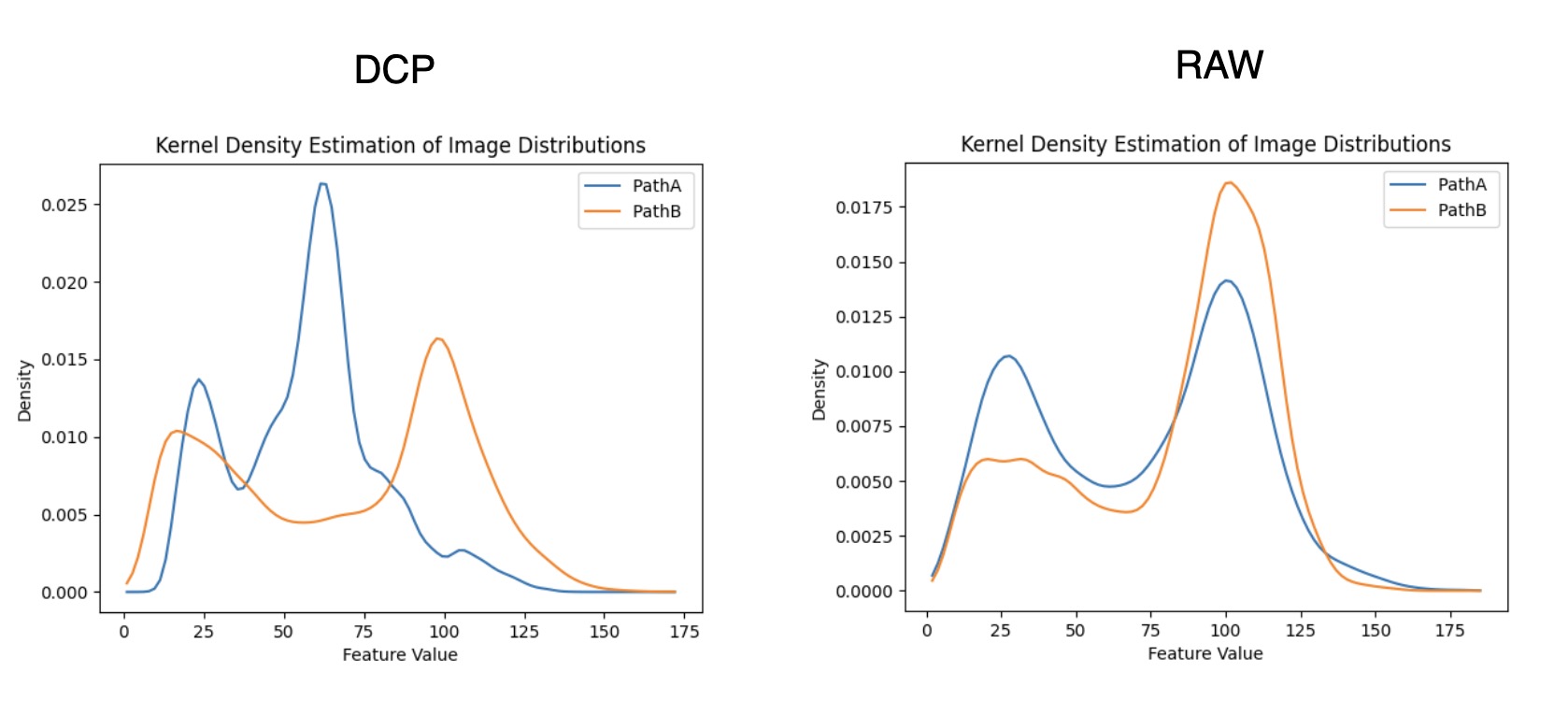}
  \caption{ The input feature distribution maps of different copy-paste path in the dual-teacher model.}
  \label{fig:distribution}
\end{figure}

\subsection{Abition study on Double-Copy-Paste method}
To validate the ability of our DCP module to bring more diversity to the dual-teacher framework and provide more guidance, we conducted relevant experiments. In Fig.\ref{fig:distribution}, we employed kernel density estimation to visualize the distribution differences in the inputs for each path within the dual-teacher model. It can be observed that DCP exhibits large distribution discrepancy between the two paths, while the raw images provide relatively minimal discrepancy information, as their distributions are comparatively more aligned. This indicates that our approach brings about more diversity. 
From Table.\ref{tab:7}, it shows that using any individual copy-paste strategy from our DCP module leads to performance improvement. However, the best results are achieved when these strategies are combined, and a noticeable performance decline is observed when the copy-paste strategies are not employed.

\begin{table}
\centering
\arrayrulecolor{black}
\caption{Ablation studies on copy-paste methods
}
\scalebox{0.8}{
\begin{tabular}{c|c|c|c|c} 
\noalign{\smallskip}
\hline
\noalign{\smallskip}
Settings                               & Dice↑ & Jaccard↑ & 95HD↓ & ASD↓  \\ 
\arrayrulecolor{black}
\noalign{\smallskip}
\hline
\noalign{\smallskip}
Only Step1 (\textcircled{\small{1}} in Fig.\ref{fig:DCP})                      & 89.62 & 81.28    & 6.87  & 1.95  \\ 
Only Step2 (\textcircled{\small{2}} in Fig.\ref{fig:DCP})                      & 89.57 & 81.18    & 7.23  & 2.15  \\ 

w/o    \textcircled{\small{1}} and \textcircled{\small{2}}                  & 86.47 & 76.50    & 12.48  & 2.71  \\

DCP method (this paper) & \textbf{90.12} & \textbf{82.04}    & \textbf{5.55}  & \textbf{1.77}  \\
\noalign{\smallskip}
\hline
\noalign{\smallskip}

\end{tabular}
}
\arrayrulecolor{black}
\label{tab:7}

\end{table}

\subsection{Abition study on Teacher model update stratedy}
To further explore the selection method for our two augmented input paths, we conducted experiments on the LA dataset with two different settings of labels, as shown in Table.\ref{tab:6}. We compared a completely random approach with switching method proposed in the previous work \cite{zhao2023alternate}, which switches teacher updates after a number of time steps. From the results, it can be observed that using the simplest completely random step achieves the best performance.
\begin{table}
\centering
\arrayrulecolor{black}
\caption{Ablation studies on path selection methods on LA dataset.
}
\scalebox{0.8}{
\begin{tabular}{c|c|c|c|c|c} 
\noalign{\smallskip}
\hline
\noalign{\smallskip}
Settings              & Label num & Dice↑          & Jaccard↑       & 95HD↓         & ASD↓           \\ 
\noalign{\smallskip}
\hline
\noalign{\smallskip}
Random path selection & 5\%labeled  & \textbf{90.12} & \textbf{82.04} & \textbf{5.55} & \textbf{1.77}  \\ 

Forced path switching & 5\%labeled  & 89.42          & 80.97          & 6.22          & 1.93           \\ 

Random path selection & 10\%labeled & \textbf{91.18} & \textbf{83.84} & \textbf{5.00} & 1.61           \\ 

Forced path switching & 10\%labeled & 90.92          & 83.43          & 5.91          & \textbf{1.57}  \\
\noalign{\smallskip}
\hline
\noalign{\smallskip}
\end{tabular}
}
\arrayrulecolor{black}
\label{tab:6}
\end{table}

\section{Conclusion}
In this paper, we propose a 3D semi-supervised medical image segmentation method based on a dual-teacher model. We improve the popular copy-paste strategy to increase the diversity of the dual-teacher models, enabling the student model to acquire more knowledge, and we design a flexible dual-teacher ensemble method to generate high-quality pseudo-labels as well. These approaches partially address the coupling issues encountered in previous research. Significant performance improvements were achieved across multiple 3D medical image datasets and in diverse settings. 

\textbf{Acknowledgement}. ~This work was supported by the NSFC Program
(62222604, 62206052, 62192783), Jiangsu Natural Science Foundation
(BK20210224), and Shandong Natural Science Foundation (ZR2023MF037).
\label{}




\bibliographystyle{elsarticle-num-names} 
\bibliography{refs}




\end{document}